\documentclass[journal]{IEEEtran}
\usepackage{cite}

\ifCLASSINFOpdf
  \usepackage[pdftex]{graphicx}
  \graphicspath{{../figures/}}
  \DeclareGraphicsExtensions{.pdf,.jpeg,.png}
\else
\fi
\usepackage{amsmath}
\usepackage{url}

\usepackage{multirow}
\usepackage{bm}
\usepackage{amssymb}
\usepackage{xspace}
\usepackage{booktabs}
\usepackage{soul}

\makeatletter
\DeclareRobustCommand\onedot{\futurelet\@let@token\@onedot}
\def\@onedot{\ifx\@let@token.\else.\null\fi\xspace}

\def\eg{\emph{e.g}\onedot} 
\def\ie{\emph{i.e}\onedot} 
 
\def\etc{\emph{etc}\onedot} 
\def\wrt{\emph{w.r.t}\onedot} 
\def\etal{\emph{et al}\onedot}
\newcommand{\figcaption}{\def\@captype{figure}\caption}
\newcommand{\tabcaption}{\def\@captype{table}\caption}
\makeatother

\begin{document}
%
\title{Narrowing the Gap: Improved Detector Training with Noisy Location Annotations}
%
%
%

\author{Shaoru~Wang,
        Jin~Gao,
        Bing~Li,
        Weiming~Hu,~\IEEEmembership{Senior Member,~IEEE}
\thanks{This work is supported by the National Key Research and Development Program of China under Grant  No. 2020AAA0140003, Natural Science Foundation of China (Grant No. 61972394, 62036011, 61721004), the Key Research Program of Frontier Sciences, CAS, Grant No. QYZDJ-SSW-JSC040. Jin Gao and Bing Li are also supported in part by the Youth Innovation Promotion Association, CAS.}
\thanks{Shaoru Wang, Jin Gao and Bing Li are with the National Laboratory of Pattern Recognition, Institute of
Automation, Chinese Academy of Sciences, Beijing 100190, China, and also
with the School of Artificial Intelligence, University of Chinese Academy of
Sciences, Beijing 100190, China (E-mail: wangshaoru2018@ia.ac.cn, \{jin.gao, bli\}@nlpr.ia.ac.cn).}
\thanks{Weiming Hu is with the CAS Center for Excellence in Brain Science and
Intelligence Technology, Chinese Academy of Sciences, Shanghai 200031,
China, also with the National Laboratory of Pattern Recognition, Institute of
Automation, Chinese Academy of Sciences, Beijing 100190, China, and also
with the School of Artificial Intelligence, University of Chinese Academy of
Sciences, Beijing 100190, China (E-mail: wmhu@nlpr.ia.ac.cn).}
}
\maketitle

\begin{abstract}
Deep learning methods require massive of annotated data for optimizing parameters. For example, datasets attached with accurate bounding box annotations are essential for modern object detection tasks. However, labeling with such pixel-wise accuracy is laborious and time-consuming, and elaborate labeling procedures are indispensable for reducing man-made noise, involving annotation review and acceptance testing. In this paper, we focus on the impact of noisy location annotations on the performance of object detection approaches and aim to, on the user side, reduce the adverse effect of the noise. First, noticeable performance degradation is experimentally observed for both one-stage and two-stage detectors when noise is introduced to the bounding box annotations. For instance, our synthesized noise results in performance decrease from 38.9\% AP to 33.6\% AP for FCOS detector on COCO test split, and 37.8\%AP to 33.7\%AP for Faster R-CNN. Second, a self-correction technique based on a Bayesian filter for prediction ensemble is proposed to better exploit the noisy location annotations following a Teacher-Student learning paradigm. Experiments for both synthesized and real-world scenarios consistently demonstrate the effectiveness of our approach, \eg, our method increases the degraded performance of the FCOS detector from 33.6\% AP to 35.6\% AP on COCO. 
\end{abstract}


%
\IEEEpeerreviewmaketitle

\section{Introduction}
%
%
%
%
\IEEEPARstart{O}{bject} detection aims at localizing individual objects and recognizing their categories. The imaginary bounding boxes are usually adopted to represent the object locations for the sake of simplicity. Considerable large-scale datasets, including PASCAL VOC~\cite{PascalVOC}, COCO~\cite{coco}, and Objects365~\cite{Objects365}, contribute to rapid advances in modern CNN-based object detection~\cite{yolo,FasterRCNN,FocalLoss,fpn,ssd,FCOS} greatly. These datasets focus on pre-defined categories, and all object instances belonging to these categories are labeled with bounding boxes together with the corresponding categories. 

\begin{figure}[t]
	\centering  
	\includegraphics[width=0.94\linewidth]{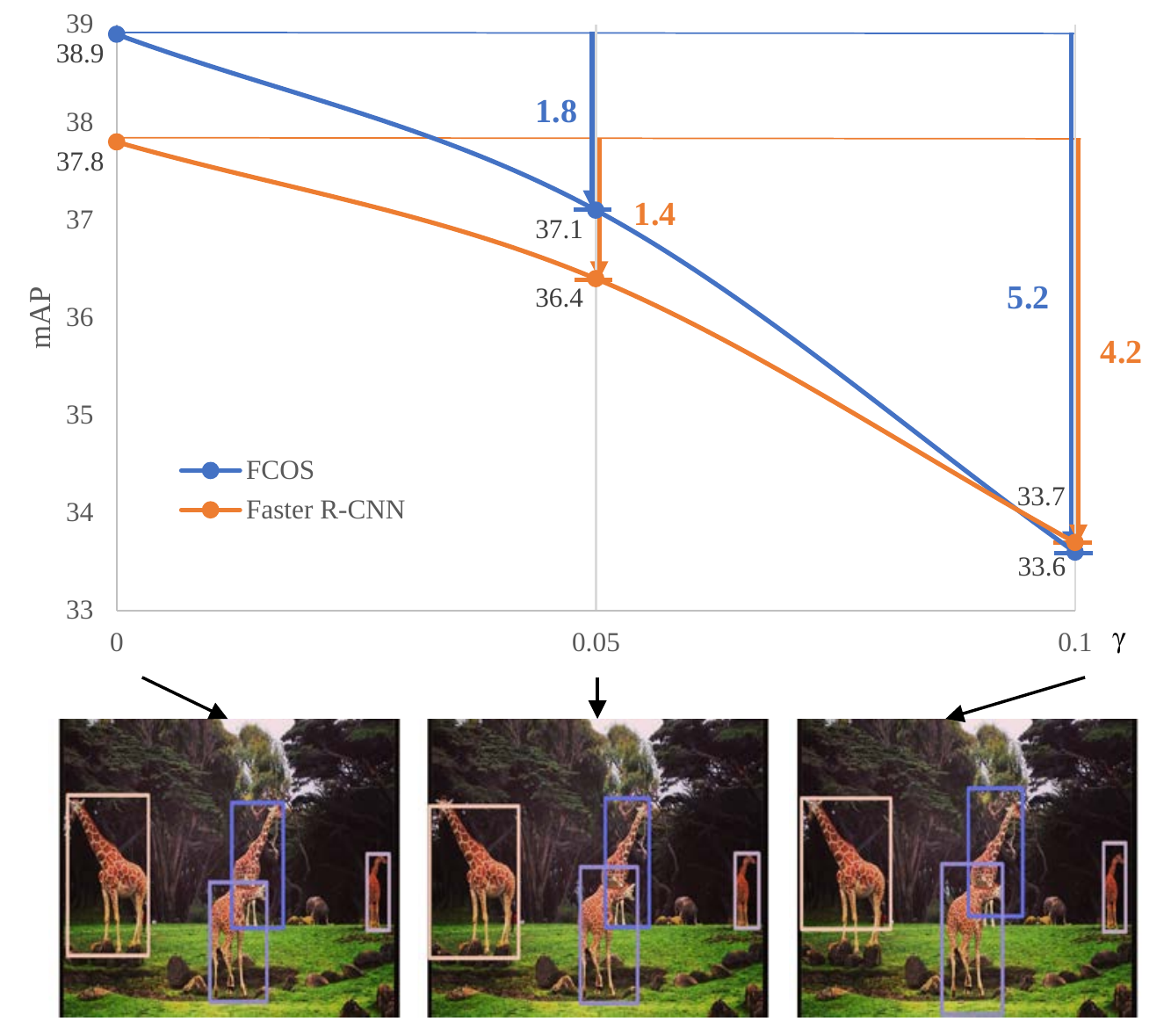}
	\caption{Noticeable degradation of performance is observed when the location annotations of the dataset used for training are corrupted by noise. The bottom figures illustrate the location annotations under different noise levels.}
	\label{fig:main}
	\vspace{-5pt}
\end{figure}

These objects' boundary coordinates are typically annotated with pixel-wise accuracy, which means annotation skills training for annotators is required before starting annotating and an explicit verification step is indispensable for achieving fine-grained annotation quality~\cite{coco}. What's more, zoom-in operations are sometimes required for annotators to watch the distinct small objects better, which makes such labor-intensive annotations even more time-consuming. So the inevitable negligence while annotating a huge dataset like COCO (which contains 143k images and 860k object annotations) may cause man-made noise involved in the annotations. Since detection has been widely applied in many different varieties of niche and emerging areas, there have been large amounts of public and private datasets for different AI-driven applications. However, not all teams can afford such huge labor or time cost and organize such a professional processing line for annotation as the COCO team has done. In some cases, one may possess a limited budget or face an impending deadline. The question is whether such fine-grained annotations like COCO are essential for the detection task, what will happen to the detectors when the annotations are corrupted by some noise due to the limited budget or impending deadline, and whether appropriate relaxation on the strict annotation standard is acceptable. Some existing studies have focused on learning with noisy category labels (\cite{reed2014training, xiao2015learning, zhang2018generalized, jiang2018mentornet} \etc), and the adverse effects on both synthesized and real-world noisy datasets have been vastly investigated. In these studies, massive effective methods are also proposed to learn robust classification models against label noise. Nevertheless, there are seldom studies focused on the noise in the location annotations for the detection tasks. How the quality of location annotations affects the performance of detectors is rarely studied.

In this paper, we first investigate the impact of noisy location annotations in the detection task, and non-negligible performance degradation is observed (see Fig.~\ref{fig:main}) when the very plain Gaussian noise is injected into the location annotations of detection datasets. We conduct all of these benchmark-simulated label noise evaluations on COCO~\cite{coco} for its popularity, which involves about 118k images for training (\texttt{train}), 5k for validation (\texttt{minival}) and 20k reserved for testing (\texttt{test-dev}). We simply add Gaussian noise to each of the box's boundary coordinates independently for each instance in the \texttt{train} split. We choose FCOS~\cite{FCOS} and Faster R-CNN~\cite{FasterRCNN}, two of the representative single-stage and two-stage detectors respectively, for evaluation. We adopt the original default training configuration of these methods, except the replacement of the annotations of the \texttt{train} split with our corrupted ones. In Fig.~\ref{fig:main}, the impact of different noise levels in terms of performance degradation is evaluated on the original COCO \texttt{test-dev} split, where 
$\gamma$ is a parameter used to control the level of the noise and the higher one indicates more severe noisy corruption. Details of the way to generate the noise are presented in Sec.~\ref{sec:analysis}. The results indicate that the detection approaches without particular treatment possibly have a poor ability to learn robust models against the noise in location annotations. We attribute it to the fact that most of these methods blindly assume the annotations to be accurate and credible. To further show the impact in the real-world scenario, we manually labeled a subset of the COCO dataset, up to 12k images, with a rather loosened label standard, contributing to about 90k rough bounding-box annotations. Performance degradation is also observed on this real-world annotation-noise subset, as shown in Sec.~\ref{sec:analysis}. Based on the above analyses, we insist that fine-grained location annotations with pixel-wise accuracy are crucial to the real-world applications of these detectors.

However, noise corruption is inevitable or incurred yet in some cases. So it is vital to study and develop effective methods to alleviate the adverse effect of the noisy location annotations in the detection task. In this paper, a simple and efficient method is proposed to make effective use of noisy annotations among large datasets. It is known that the ensemble methods~\cite{ensemble} may have the ability to integrate multiple predictions from different models and get closer to the real noise-free labels. We extend this idea to the ensemble of predictions from an individual detector. To this end, an ensemble method based on a Bayesian filter is proposed, which can correct the noisy box annotations by the predictions of the detectors themselves. This integration considers both the probability of the prediction belonging to the corresponding object and its classification confidence. Furthermore, to avoid the trivial solution during the direct use of this self-correction process in training, a Teacher-Student learning framework is adopted. Specifically, a teacher detector is first trained on the noisy dataset and then used to correct box coordinate noise by prediction ensemble. The student detector is then trained based on the corrected annotations and we achieve better performance as expected in comparison with the frozen teacher detector. In other words, we boost the student detector's locating performance by exploiting the self-correction ability of the noisy teacher detector.

In summary, our contributions are three-fold:

\begin{itemize}
    \item Some insightful analyses on the impact of noisy location annotations for the detection task are presented and noticeable performance degradation is observed in both synthesized and real-world scenarios. 
    \item A novel prediction ensemble approach based on a Bayesian filter is proposed, which enables better exploitation of noisy annotations in a Teacher-Student learning framework.
    \item Our method provides a simple baseline for the study of noisy location annotations, which is expected to advance the research related to localization tasks, \eg, object detection, tracking, \etc.
\end{itemize}

The code and our newly-annotated noisy subset will be made publicly available at \url{https://github.com/wangsr126/NDet} to promote the study on this topic.

\section{Related Works}

\subsection{Object Detection}
Modern deep-learning-based methods for object detection can be generally formulated as multi-region classification and regression tasks. First, a large number of anchors are tiled on the image, and each of them is assigned to one of the ground-truth objects in the image or just the background. Then the category and relative coordinates to the specified nearby ground-truth object (if exists) are predicted for each anchor~\cite{FasterRCNN, FocalLoss, fpn}. A class of methods cascades the above procedures several times to refine predictions, which are commonly known as \textit{two-stage} methods~\cite{FasterRCNN, CascadeRCNN}. Another interesting direction simplifies the anchor from the box to a single point, leading to a simpler paradigm and more potential in terms of generalization ability~\cite{FCOS, CenterNet}, namely \textit{anchor-free} methods. In this paper, we focus on the regression branch for the above detectors. Relative coordinates from the anchor\footnote{Anchor box and anchor point are collectively referred to as anchor in the following part without loss of generality.} to the ground-truth box of the assigned object are predicted in the regression branch. The $L_p$-norm-based loss (\eg, smooth $L_1$ loss~\cite{FasterRCNN}, $L_2$ loss~\cite{yolo, yolov3}, $L_1$ loss~\cite{mmdetection}) and IoU-based loss (\eg, IoU loss~\cite{unitbox}, GIoU loss~\cite{giou}) are usually adopted for supervision. However, these methods assume the annotations of location are absolutely accurate, without modeling the noise in the annotations, which shows poor anti-noise performance. KL-loss~\cite{kl-loss} considers the inherent ambiguities of the ground-truth bounding boxes, but it also assumes the annotations are accurate for most instances, which is thus orthogonal to our case.

\subsection{Label Noise}\label{sec:label-noise}
Corrupted labels are ubiquitous in many real-world collected and annotated datasets, and can severely impair the performance of deep neural networks especially when the dataset is annotated with a limited budget. The label noise may arise upon mistakes of human annotators or automatic label extraction tools. For classification tasks, several approaches have sought to promote the robustness of classifiers to label corruption. One representative class of methods aims to reduce noisy samples' impact with carefully designed losses~\cite{zhang2018generalized, arazo2019unsupervised, ghosh2017robust}, regularization terms~\cite{li2019learning, Menon2020Can} or adaptive sample re-weighting~\cite{jiang2018mentornet, ren2018learning,decoupling, coteaching}. Another direction focuses on improving the label quality by correcting the noisy labels~\cite{reed2014training, xiao2015learning, tanaka2018joint, li2017learning}. Most of the above studies usually conduct experiments on noisy datasets generated from CIFAR-10, CIFAR-100 by artificially corrupting the true labels. Few of them~\cite{zhang2018generalized, arazo2019unsupervised} conduct extensive experiments on real-world noisy datasets for classification tasks.

Departing from them, our approach mainly focuses on correcting the noisy location annotations in object detection tasks, which bears some similarities to the above trend of correcting noisy labels. Some of the methods designed for classification tasks can be naturally extended to detection tasks~\cite{reed2014training, vahdat2017toward} to solve noisy category labels, \eg, mistaken labels or missing labels. But little attention has been paid to the noise in location annotations. Some works for detection focus on semi-supervised settings, where only a subset of data are annotated while others remain unlabeled~\cite{jeong2019consistency, sohn2020simple, liu2021unbiased}, or weakly-supervised settings, where only image-level category tags are available~\cite{Bilen_2016_CVPR, tang2017multiple, Tang_2018_ECCV, rosenberg2005semi}. All of these paradigms show significant differences to our settings. The most related works to ours are the noisy location annotation study conducted by Mao \etal~\cite{mao2020noisy} and some followers~\cite{chadwick2019training, mao2021noisy}.
In contrast to these works, we additionally analyze the pattern of human-made label noise and the generalization performance to the real-world noisy dataset, which shows more practical significance.

\subsection{Teacher-Student Learning}
Teacher-Student learning is a learning paradigm that uses a teacher network to train a student network for various vision tasks. The idea is introduced in knowledge distillation~\cite{hinton2015distilling}, in which a teacher model is generally pre-trained on a specific task and kept frozen during distillation. Knowledge distillation is widely adopted in various tasks, \eg, classification~\cite{hinton2015distilling, heo2019comprehensive, tung2019similarity, you2017learning}, detection~\cite{chen2017ODKD, li2017mimicking}, and segmentation~\cite{he2019knowledge, liu2019structured}. The Teacher-Student learning framework is also adopted in semi-supervised learning~\cite{liu2021unbiased, mean-teacher, sohn2020fixmatch}. For instance, Mean Teacher~\cite{mean-teacher} constructs the teacher network as a moving average of student network and the predictions of the teacher are seen as pseudo labels for the student to learn against.  Besides, it is also widely used in self-supervised learning~\cite{caron2021emerging}, domain adaptation~\cite{8461682}, \etc Our Teacher-Student framework bears some similarities to knowledge distillation. However, the improvement of the student in distillation comes from the additional knowledge of the teacher, while the teacher itself in our method may not show superiority to the student. The improvement in our method is derived from our newly-designed annotation correction module, which can serve as a better supervision for the student detector. Hopefully, our framework is compatible with other distillation methods, though it is beyond the scope of this paper.


\section{Methodology}
In this section, we first briefly describe the modern detectors as preliminary. Then some experimental analyses on the impact of noisy location annotations are conducted. Finally, a simple yet effective self-correction technique is proposed based on the prediction ensemble paradigm, which can effectively improve the quality of location annotations. Along with the Teacher-Student learning framework, our method can practically narrow the performance gap between detectors with and without noisy location annotations. 

\subsection{Preliminary on Object Detection}
Most modern detectors~\cite{FasterRCNN, FocalLoss, fpn, FCOS} follow a dense prediction paradigm. First, a large number of anchors are tiled on the image. During the training phase, each of them is assigned to one of the objects in the image or just to the background, and fancy losses are designed to guide them to predict the right category and regress the relative coordinates to the specified nearby ground-truth object (if exists). The above process can be formulated as follows. 

For each anchor $a_i$, classification scores $\bm{p_i}$ and parameterized coordinates $\bm{t_i}$ are predicted, where, $\bm{p_i}$ is usually a $C$-dimension vector, representing probabilities of belonging to the predefined $C$ categories, and $\bm{t_i}$ is a $4$-dimension vector, usually representing the relative offsets from the anchor to the predicted bounding box. Then $a_i$ is assigned to one of the objects in the image or just to the background, consequently constructing the class label $c_i^*$ and regression target $\bm{t}_i^*$. The latter is encoded based on the assigned ground truth box $\bm{b}_j^*$, and $\bm{b}_j^*=[l_j^*, r_j^*, t_j^*, b_j^*]$ for four object bound coordinates.
The following training loss function is applied on these predictions:

\begin{equation}\label{equ:loss}
\begin{aligned}
    L(\{\bm{p}_i\}, \{\bm{t}_i\})&=\lambda_{cls}\sum_{i}L_{cls}(\bm{p}_i, c_i^*)\\
    &+\lambda_{reg}\sum_{i}\mathbb{I}_{\{c_i^*>0\}}L_{reg}(\bm{t}_i, \bm{t}_i^*),
\end{aligned}
\end{equation}
where $L_{cls}$ is for category classification task, and cross entropy loss~\cite{FasterRCNN} and focal loss~\cite{FocalLoss} are usually adopted. $L_{reg}$ is for bounding box regression task and smooth $L_1$ loss~\cite{FasterRCNN}, $L_2$ loss~\cite{yolo, yolov3}, $L_1$ loss~\cite{mmdetection}, IoU loss~\cite{unitbox}, GIoU loss~\cite{giou}, \etc, can be adopted. 
$\mathbb{I}_{\{c_i^*>0\}}$ is the indicator function, being 1 if $c_i^*>0$ (foreground) and 0 otherwise (background). $\lambda_{cls}$ and $\lambda_{reg}$ are both the factors used for normalization and re-weighting simultaneously. 
The above procedure can be cascaded several times to further refine the predictions through regarding the predicted boxes from the previous stage as the anchors for the current stage~\cite{FasterRCNN, CascadeRCNN}. 

During the inference phase, $\bm{p}_i$ and $\bm{t}_i$ are predicted for each anchor $a_i$, and non-maximum suppression (NMS) is applied as post-processing to remove the redundant predictions at last.

\begin{figure*}[t]
	\centering  
	\includegraphics[width=0.9\linewidth]{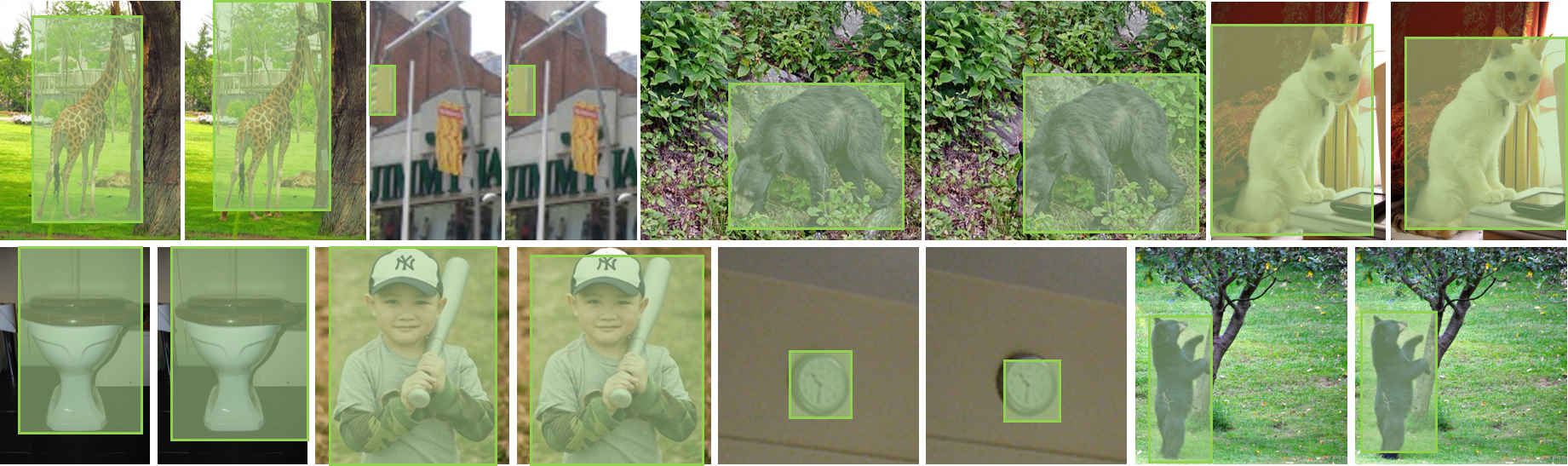}
	\caption{Examples of our newly labeled images from the COCO \texttt{train} split. In each pair, the left one is the original annotation, and the right one is ours.}
	\label{fig:example}
	\vspace{-5pt}
\end{figure*}

\begin{figure}[t]
	\centering  
	\includegraphics[width=1\linewidth]{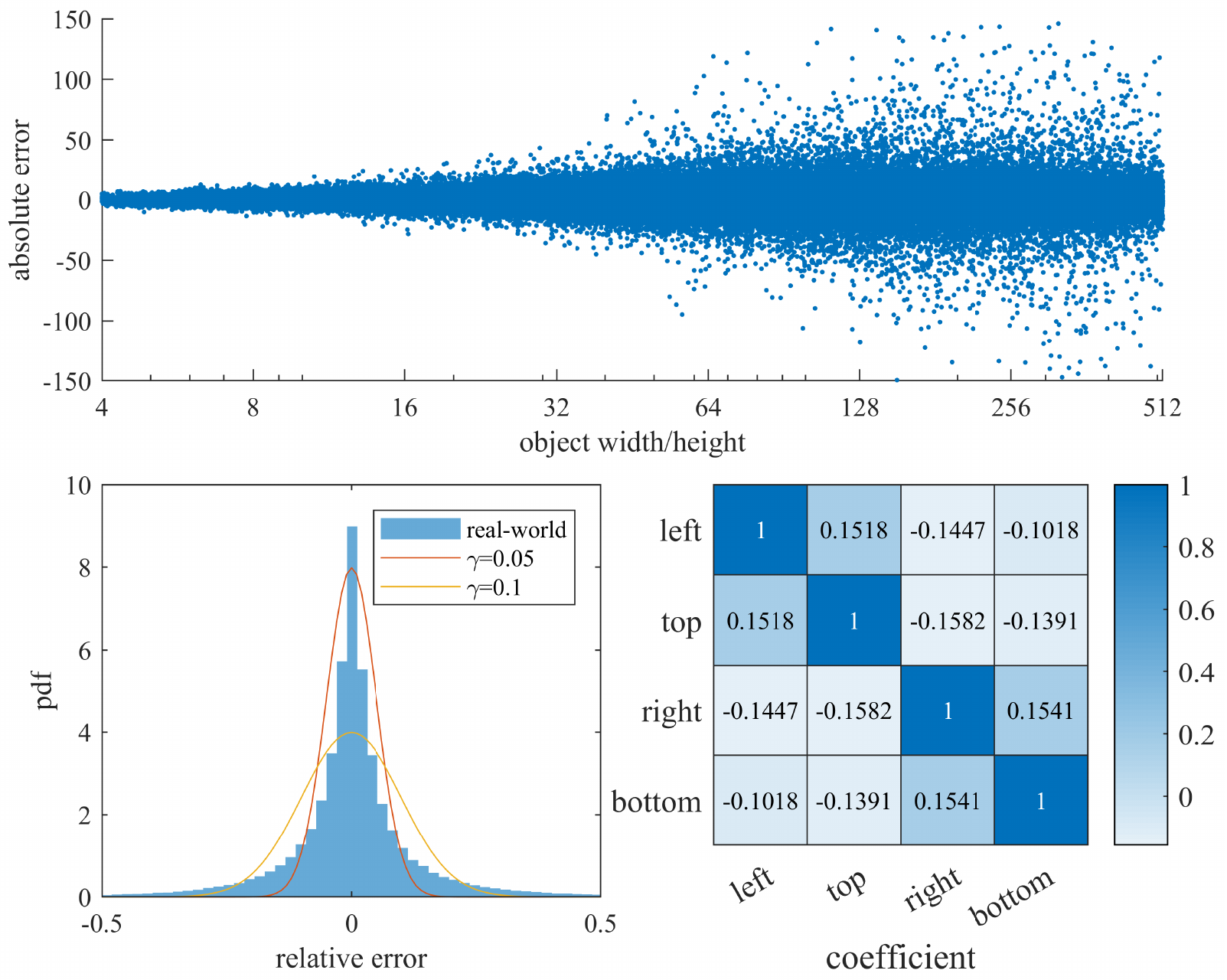}
	\caption{Analyses of man-made noise in location annotations. Top: scatter diagram for absolute error of each boundary annotation with respect to corresponding object width or height. Bottom left: the distribution of relative error; the solid curves illustrate the distributions of our synthesized ones. Bottom right: correlation coefficients between relative errors of different boundary coordinates.}
	\label{fig:noise}
	\vspace{-5pt}
\end{figure}

\subsection{Analyses on the Impact of Noise}\label{sec:analysis}

In most previous works, the location annotations in datasets for evaluation are, most commonly and tacitly, treated as accurate ground truth labels without considering the possible man-made errors, which means the regression target $\bm{t}_i^*$ in Equation~\eqref{equ:loss} is considered to be accurate. Such conditions can be satisfied for some publicly available datasets thanks to the elaborate labeling effort~\cite{coco, PascalVOC, Objects365}, whereas they are usually not held for some private and self-collected datasets. 
If the location annotations are not accurate enough due to the limited budget, how much is the impact on the performance of modern detectors? To answer this question, experiments for both synthesized and real-world scenarios are conducted in this paper. 

At the beginning of our analyses, we vastly labeled a subset of COCO \texttt{train} split, up to 12k images with relaxed label restrictions. The only request for annotators is that they should make it possible for others to retrieve the right object according to their labeled bounding box. Besides, we provide annotators with another indicator from COCO annotations to ensure our new annotations share almost the same mistaken and missing errors to COCO. In this way, the only difference between COCO annotations and ours is that the location annotations from COCO are relatively fine-grained and accurate while ours are coarse and noisy. Some examples are shown in Fig.~\ref{fig:example}. We further analyze the pattern of the man-made noise in the location annotations by considering COCO as the clean ones, as shown in Fig.~\ref{fig:noise}. We mainly have three observations as follows: 1) the variance of the noise is roughly linear to the object scale; 2) relative errors (obtained by normalizing the absolute coordinate errors with respect to corresponding object widths or heights) are mostly centered around 0; 3) the correlation coefficients between relative errors of different boundary coordinates are at a relatively low level.

\begin{figure}[t]
	\centering  
	\includegraphics[width=0.75\linewidth]{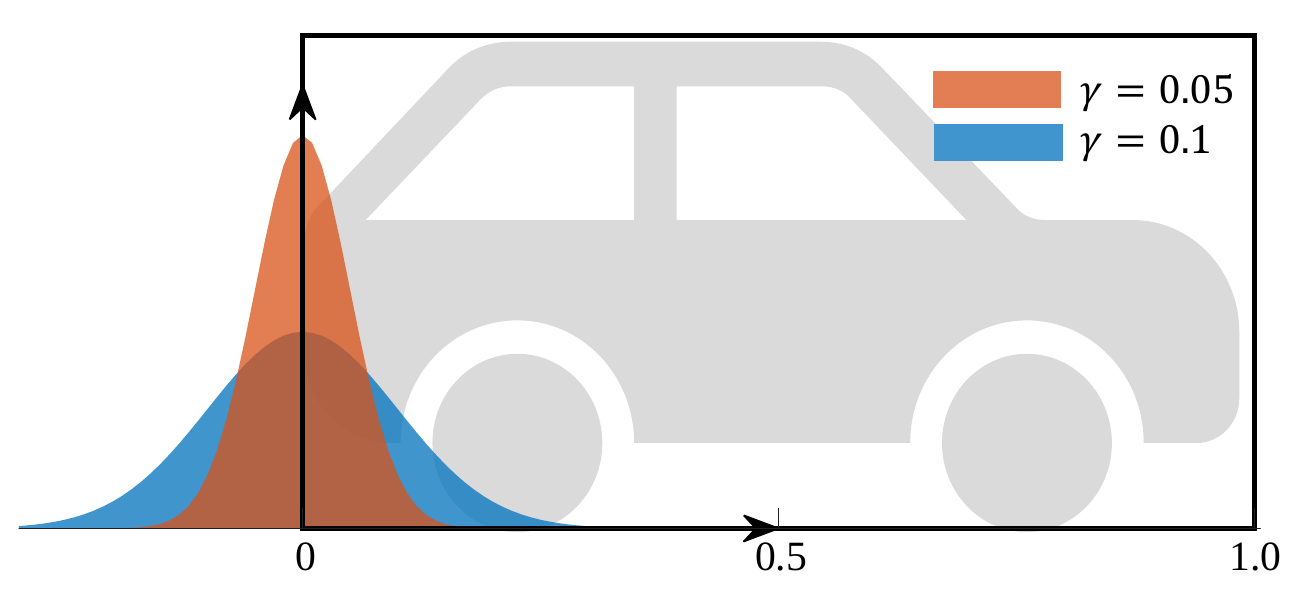}
	\caption{Noise distributions we adopt for synthesized location annotations.}
	\label{fig:gaussian}
	\vspace{-5pt}
\end{figure}

\begin{table*}[h]
\begin{minipage}[b]{.62\linewidth}
\centering
    \caption{Noticeable degradation of performance on the COCO \texttt{minival} split is observed when the location annotations of the \texttt{train} split are corrupted by synthesized and real-world noise. }
    \resizebox{1.0\linewidth}{!}{
    \renewcommand{\arraystretch}{0.8} 
    \begin{tabular}{c|c|c|c|c|c c|c c c}
        \toprule
        Methods & type & scale & $\gamma$ & $AP$ & $AP_{50}$ & $AP_{75}$ & $AP_s$ & $AP_m$ & $AP_l$ \\
        \midrule
        \multirow{5}{*}{FCOS~\cite{FCOS}} & \multirow{3}{*}{synthesized} & \multirow{3}{*}{123k} & 0 & \textbf{38.5} & \textbf{57.3} & \textbf{41.6} & \textbf{22.5} & \textbf{42.4} & \textbf{49.7} \\
         & & & 0.05 & 37.1 (-1.4) & 56.9 & 40.3 & 21.6 & 40.4 & 47.8 \\
         & & & 0.1 & 33.5 (-5.3) & 54.9 & 36.0 & 19.6 & 37.0 & 43.1 \\
        \cmidrule{2-10}
         & \multirow{2}{*}{real-world} & \multirow{2}{*}{12k} & 0 & \textbf{22.8} & \textbf{39.2} & \textbf{23.3} & \textbf{11.6} & \textbf{24.8} & \textbf{30.0} \\
         & & & - & 21.2 (-1.6) & 38.4 & 20.7 & 10.3 & 22.9 & 29.4 \\
        \midrule
        \multirow{5}{*}{Faster R-CNN~\cite{FasterRCNN}} & \multirow{3}{*}{synthesized} & \multirow{3}{*}{123k} & 0 & \textbf{37.5} & \textbf{58.2} & \textbf{40.9} & \textbf{21.1} & \textbf{41.2} & \textbf{49.0} \\
         & & & 0.05 & 35.8 (-1.7) & 57.8 & 39.0 & 20.9 & 39.3 & 45.8 \\
         & & & 0.1 & 33.3 (-4.2) & 56.3 & 35.6 & 19.3 & 37.5 & 42.1 \\
        \cmidrule{2-10}
         & \multirow{2}{*}{real-world} & \multirow{2}{*}{12k} & 0 & \textbf{23.3} & \textbf{42.9} & \textbf{22.7} & \textbf{11.7} & \textbf{25.8} & \textbf{30.8} \\
         & & & - & 21.8 (-1.5) & 42.2 & 20.2 & 10.2 & 24.4 & 29.8 \\
        \bottomrule
    \end{tabular}
    }
    \label{tab:main}
    \vspace{-18pt}
\end{minipage}
\hfill
\begin{minipage}[b]{.36\linewidth}
\centering
\includegraphics[width=0.9\linewidth]{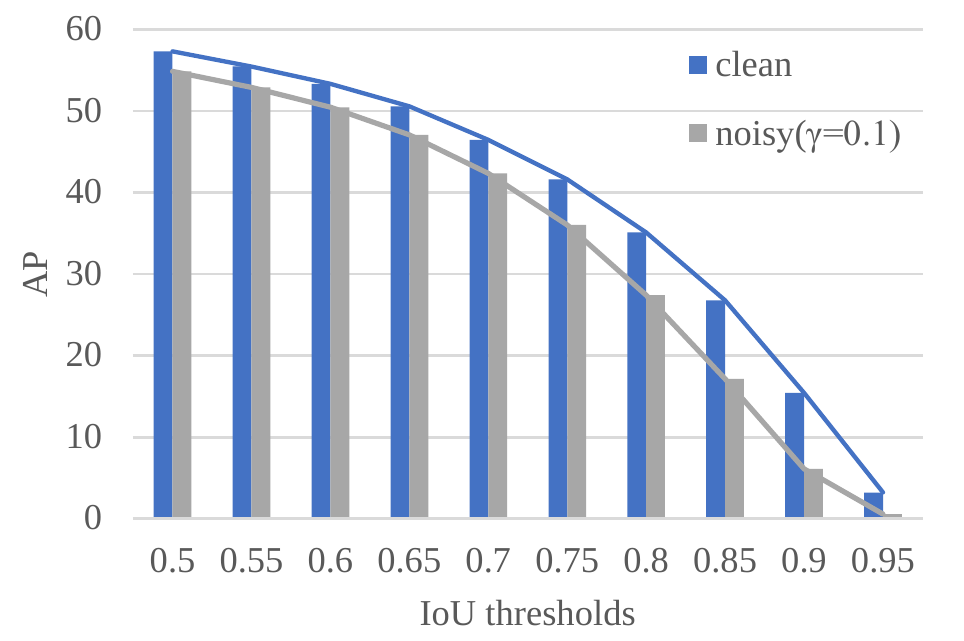}
\vspace{-5pt}
\figcaption{APs under various IoU thresholds for the FCOS detectors trained with the clean and synthesized noisy annotations.}
\label{fig:AP}
\end{minipage}
\vspace{-5pt}
\end{table*}

Despite that we have labeled such a noisy subset in the sense of real-world noisy conditions, its small scale (\ie, only 10\% of COCO dataset) and lack of ability to manually control the level of noise hold us back to deliver a deeper analysis. So we manually corrupt the location annotations by synthesized noise, following the above observations. Specifically, we add Gaussian noise to each box boundary coordinate separately for each instance in the \texttt{train} split. Take the left boundary (denoted as $l^*$) for example, the new coordinate could be $\tilde{l^*} = l^* + N_{l^*}$. The noise is set to be zero-mean with standard deviation linear to the object scale, \ie, $N_{l^*}\sim \mathcal{N}(0, \sigma_{l^*}^2)$, where $\mathcal{N}$ denotes Gaussian distribution. We introduce the root mean squared relative error as the metric to measure the level of the noise, which is defined as, \eg, $\gamma=\sqrt{\mathrm{E}[(N_{l^*}/w^*)^2]}$ for the left boundary. Thus, the level of the above Gaussian noise can be measured as $\gamma=\sigma_{l^*}/w^*$, where $w^*$ is the object width. Illustrations are shown in Fig.~\ref{fig:gaussian}. We let $\gamma$ shared across four boundaries, \ie, $\sigma_{r^*}=\gamma w^*$, $\sigma_{t^*}=\gamma h^*$, $\sigma_{b^*}=\gamma h^*$, where $\sigma_{r^*}, \sigma_{t^*}, \sigma_{b^*}$ are the standard deviations for the noise on the other three boundaries and $h^*$ is the object height.

Based on our newly labeled and synthesized annotations on the COCO dataset, we take FCOS~\cite{FCOS} and Faster R-CNN~\cite{FasterRCNN}, two of the representative single-stage and two-stage detectors respectively, as examples to examine the impact of the noise. FCOS is an anchor-free detector\footnote{We adopt the improved version with tricks like \texttt{norm\_on\_bbox}, \texttt{centerness\_on\_reg}, \texttt{center\_sampling} and \texttt{giou} unless otherwise stated.} with IoU-based regression loss, and the detector of Faster R-CNN utilizes \textit{ROIAlign} operator to formulate a two-stage framework with $L_p$-norm-based regression loss. As shown in Fig.~\ref{fig:main} and Table~\ref{tab:main}, both the two methods are influenced by the injected noise. When $\gamma=0.05$ in the case of synthesized noisy annotations, mAPs evaluated on the \texttt{minival} split degrade to 96.3\% and 95.5\% of the original scores for FCOS and Faster R-CNN respectively. We further increase $\gamma$ to $0.1$, and the performances decline to 87.0\% and 88.8\% of the original scores disastrously. Besides, we can find that APs with higher IoU thresholds drop heavier, as shown in Fig.~\ref{fig:AP}, which indicates that the noise prominently damages the ability of detectors to accurately locate objects.

\begin{figure}[t]
	\centering  
	\includegraphics[width=1\linewidth]{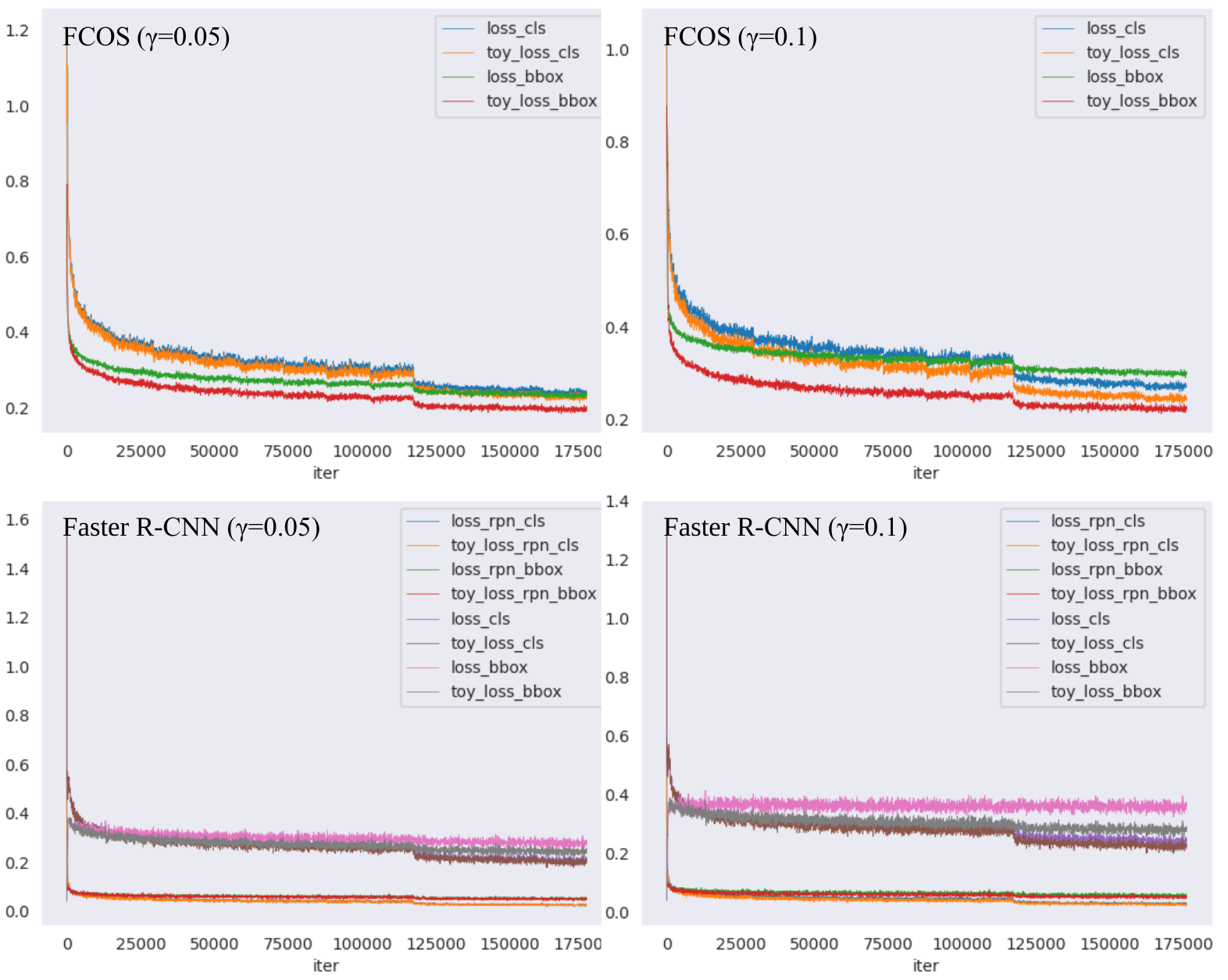}
	\vspace{-5pt}
	\caption{Training curves for detectors with original and toy training losses.}
	\label{fig:plot_loss}
	\vspace{-5pt}
\end{figure}

We also examine whether the detectors can overfit the noisy annotations. Since we have clean COCO annotations, it is easy and straightforward to monitor additional toy losses calculated by the predictions and the clean COCO annotations but stop the backward pass for computing gradients from these losses during the training phase. 
As shown in Fig.~\ref{fig:plot_loss}, we can find that the losses calculated by the noisy annotations are always higher than those calculated by the clean ones, and all of them consistently show a downward trend during training. Thus we infer it is hard to overfit for detection tasks in our settings.

\subsection{Our Improved Detector Training with Noise}
When we happen to have a disappointing dataset with noisy location annotations due to the limited labor or time budget, it is thus worthy to study how to exploit these noisy annotations effectively as direct utilization of them results in prominent performance degradation as shown in Sec.~\ref{sec:analysis}. In this section, a simple yet effective method is proposed based on the prediction ensemble to correct the noisy box annotations. Furthermore, to avoid the trivial solution during the direct use of this self-correction process in tracing, a Teacher-Student learning framework is adopted. Our method effectively alleviates the adverse effect of noisy annotations and achieves better performance on noisy datasets.

\subsubsection{Prediction Ensemble}\label{sec:correct}

Due to the popular one-to-many label assignment during detector training~\cite{FasterRCNN, FocalLoss, FCOS}, there are usually duplicate predictions corresponding to the same instance. We believe that these predictions possess the potential of correcting the noisy box annotations. To this end, we present the following problem: \textit{For one object in the image with category label $c^*$ and noisy box annotation $\bm{b}^*$, given multiple predictions $\mathcal{O}=\{O_1, O_2, \dots, O_N\}$, how to acquire the corrected bounding box $\bm{b}^{cor}$ for that object?}

Here $O_i$ is the prediction of the anchor $a_i$, and usually involves the classification score $\bm{p}_i$ and the box coordinate prediction $\bm{b}_i$.
Note that $\bm{b}_i$ denotes the absolute coordinates in the image plane, which is the decoded box prediction transformed from $\bm{t}_i$, and $N$ is the total number of anchors in the image. 

The above problem can be formulated as the estimation of the posterior probability distribution, $p(\bm{b}^{cor}|c^*,\bm{b}^*, \mathcal{O})$. 
If we cast the coordinate prediction $\bm{b}_i$ in $O_i$ as one measurement of $\bm{b}^{cor}$, and assume the measurements conditioned on $\bm{b}^{cor}$ are independently distributed, this problem can be considered as a particular case of recursive Bayesian estimation problem on \textit{static systems}. If we further assume the measurement noise is normally distributed, it can be solved by Kalman filter~\cite{Kalman-filter}. In this \textit{static system}, the output state vector for representing the corrected bounding box is updated by the above measurements that are introduced in a sequential fashion without considering their orders. This process is governed by the following equations:

\begin{equation}
    \begin{aligned}
    \bm{b}^{cor}_k&=\bm{b}^{cor}_{k-1},\\
    \bm{b}_k&=\bm{b}^{cor}_k+\bm{v}_k,
    \end{aligned}
\end{equation}
where $p(\bm{v}_k) \sim \mathcal{N}(0, R_k)$ and $R_k$ is the covariance matrix for the $k$-th measurement. This formulation can also be represented as $p(\bm{b}_k|\bm{b}^{cor}_k) \sim \mathcal{N}(\bm{b}^{cor}_k, R_k)$ in a probabilistic manner. The initial state can be modeled by the given noisy annotations, \ie, $p(\bm{b}^{cor}_0)\sim \mathcal{N}(\bm{b}^*, P^*)$, where $P^*$ is the covariance matrix for the annotations.

According to the derivations in Kalman filter~\cite{Kalman-filter}, we can obtain the following recursive equations:
 
\begin{equation}\label{equ:multi-d}
    \begin{aligned}
    K_{k}&=P_{k-1}(P_{k-1}+R_k)^{-1},\\
    \bm{b}^{cor}_k&=\bm{b}^{cor}_{k-1}+K_{k}(\bm{b}_k-\bm{b}^{cor}_{k-1}),\\
    P_k&=(I-K_k)P_{k-1},
    \end{aligned}
\end{equation}
where $K_k$ is the Kalman gain and $I$ is the identity matrix. Besides, we have the initial state as $\bm{b}^{cor}_0=\bm{b}^*, P_0=P^*$.

By some derivations, the above recursive equations can be simplified to the following general terms, which can directly estimate the coordinates based on the noisy annotation and $k$ predictions:
\begin{align}
\bm{b}^{cor}_k&=P_{k}\Big((P^*)^{-1}\bm{b}^*+\sum_{i=1}^kR_{i}^{-1}\bm{b}_i\Big),\\
P_k^{-1}&=(P^*)^{-1}+\sum_{i=1}^kR_{i}^{-1}.
\end{align}
It is shown that the order of input sequence of predicted bounding boxes \emph{does not} matter in estimating $\bm{b}^{cor}_k$ and $P_k$. More specifically, we can use all the $N$ predictions to estimate the corrected coordinates as follows:
\begin{equation}\label{equ:final}
    \bm{b}^{cor}=\bm{b}^{cor}_N=(I+\sum_{i=1}^NP^*R_i^{-1})^{-1}(\bm{b}^*+\sum_{i=1}^NP^*R_i^{-1}\bm{b}^{cor}_i).
\end{equation}

It can be drawn that $P^*R_i^{-1}$ matters in Equation~{\eqref{equ:final}} rather than any individual one of $P^*$ and $R_i$. As the measurements (or predictions) are also trained with the supervision from the noisy annotations,
we thus assume that the covariance matrix $R_i$ for the $i$-th measurement inherits the correlationship in the covariance matrix $P^*$ for the annotation and follows a scaling relation, \ie, $P^*=\delta_i R_i$, where the scalar $\delta_i$ is the scale factor between $P^*$ and $R_i$. Note that $P^*$ and $R_i$ are not necessarily to be diagonal matrices, which means that the four coordinates of the box are not necessarily independent to each other. Considering most objects in COCO~\cite{coco} are subject to certain length/width ratios, they are indeed not independent. Following the assumption that $P^*=\delta_i R_i$, the Equation~{\eqref{equ:final}} can be rearranged to the one-dimensional case for each coordinate of the box. Here we take the estimation of the left boundary coordinate ($l^{cor}$) for example:
\begin{equation}
    l^{cor}=\frac{l^*+\sum_{i=1}^N \delta_i l_i}{1+\sum_{i=1}^{N}\delta_i},
\end{equation}
where $l^*$ and $l_i$ are the left boundary coordinate of the annotation and the $i$-th measurement.

Then, we discuss the design of $\delta_i$. At least two factors should be considered: 1) the probability of the prediction's bounding box fitting the corresponding object; 2) the confidence of the prediction's category. To this end, we specially design $\delta_i$ as follows:
\begin{figure}[t]
	\centering  
	\includegraphics[width=0.8\linewidth]{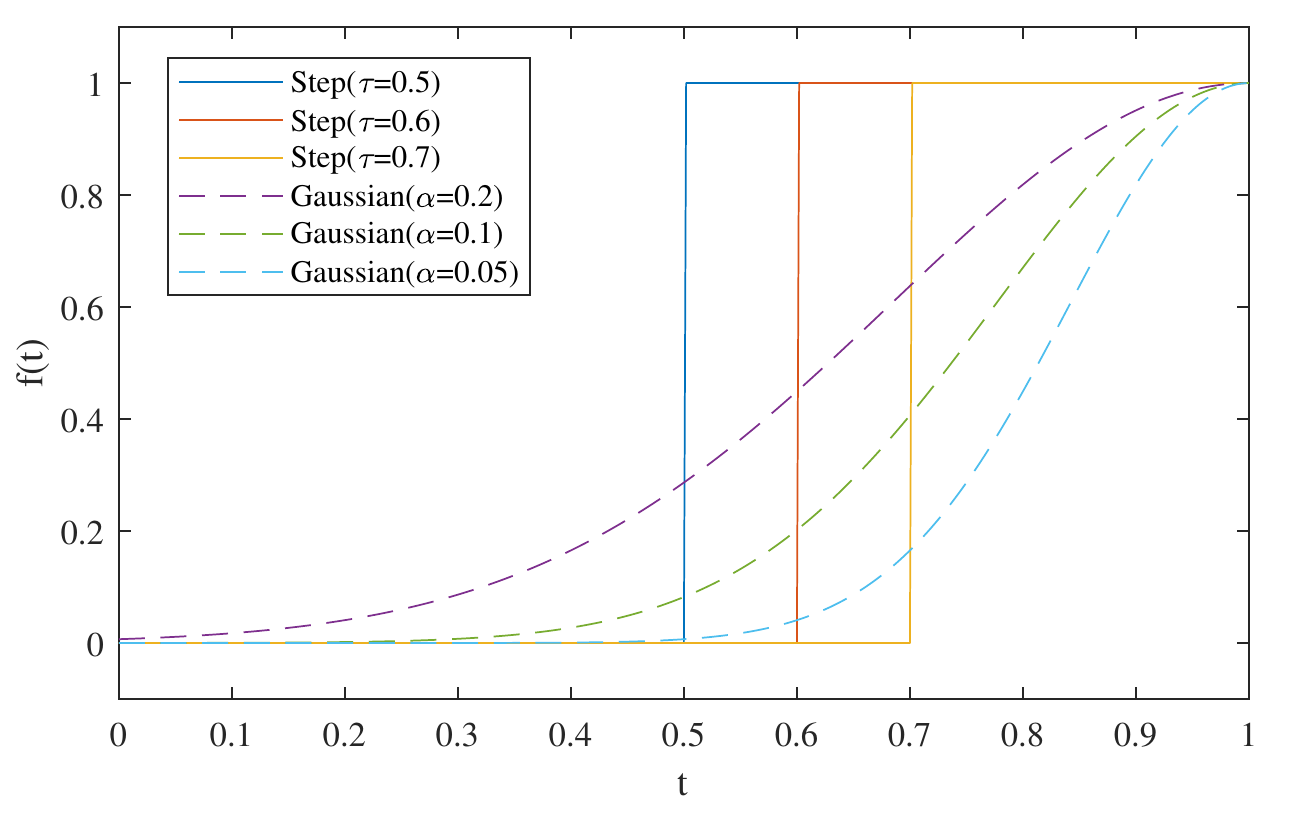}
	\vspace{-5pt}
	\caption{Various choices for $f(t)$ in Equation~\eqref{equ:delta}: \texttt{Step} for $f(t)=\mathbb{I}(t\geqslant \tau)$ and \texttt{Gaussian} for  $f(t)=e^{-\frac{(1-t)^2}{\alpha}}$.}
	\label{fig:f}
	\vspace{-5pt}
\end{figure}

\begin{figure*}[t]
    \begin{minipage}[b]{.63\linewidth}
	\centering  
	\includegraphics[width=1.0\linewidth]{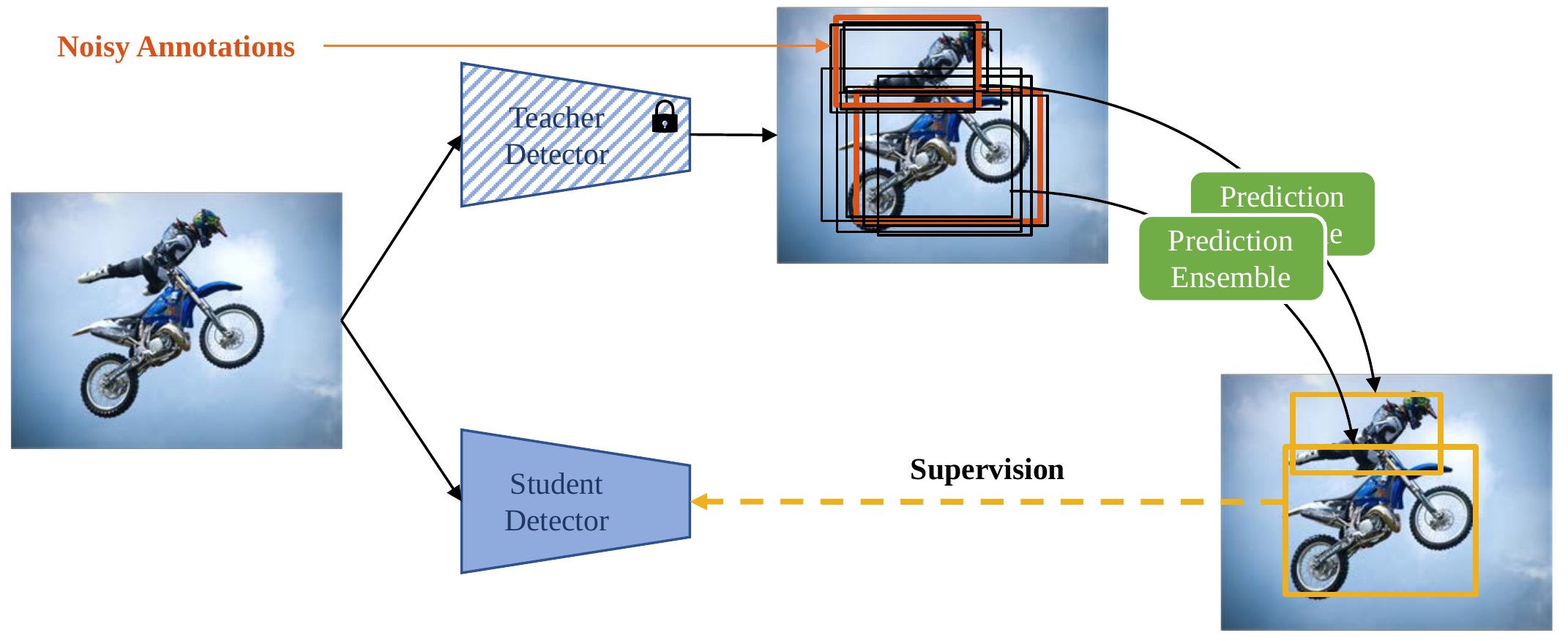}
	\vspace{-5pt}
	\caption{The whole Teacher-Student learning framework. A frozen teacher detector and another student detector process images simultaneously. The predictions of the teacher detector are then utilized to correct the noisy bounding box annotations using our self-correction method described in \ref{sec:correct}. The corrected annotations are used as supervision for the student detector.}
	\label{fig:structure}
	\vspace{-10pt}
	\end{minipage}\hfill
    \begin{minipage}[b]{0.36\linewidth}
	\centering  
	\includegraphics[width=0.9\linewidth]{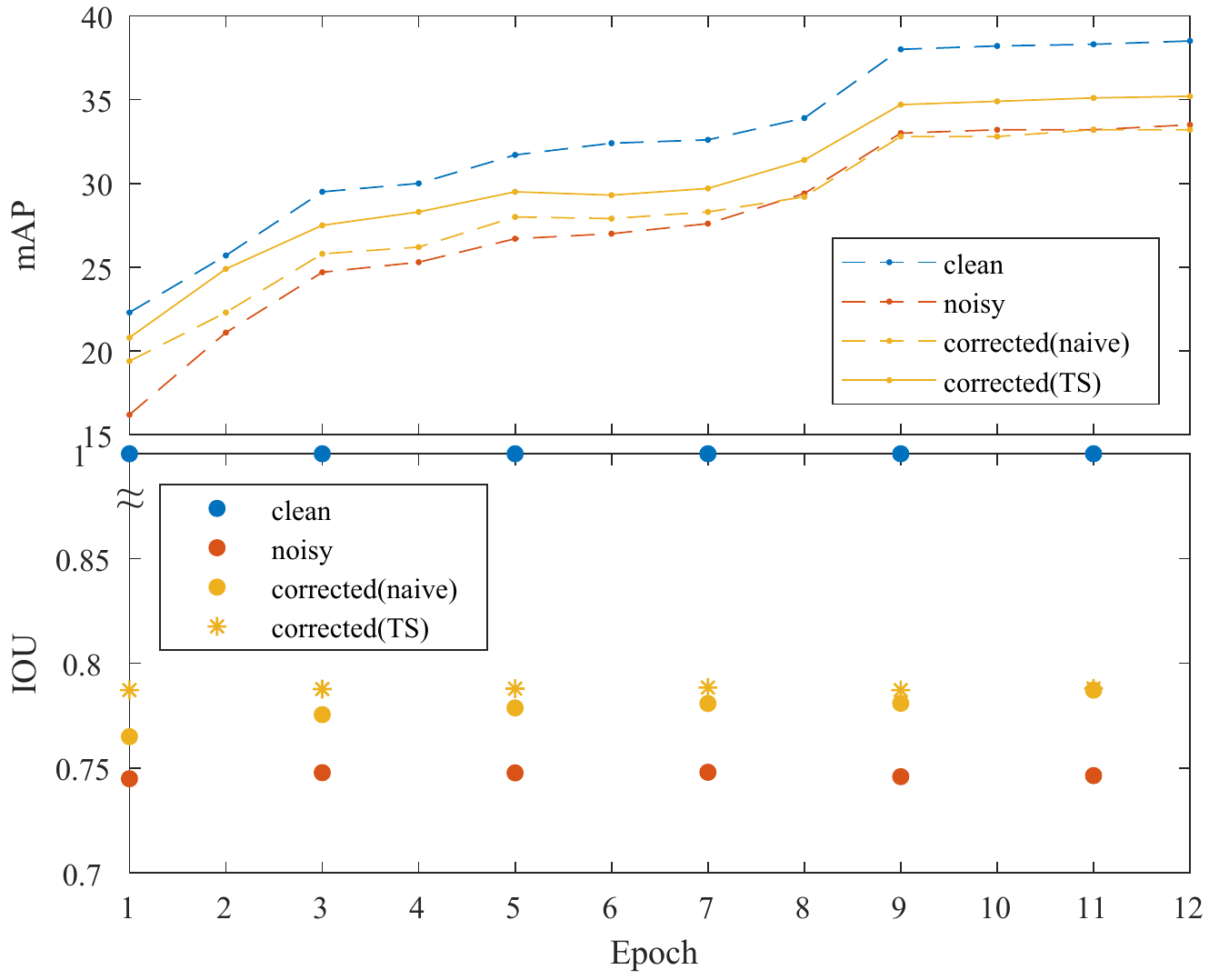}
	\vspace{-10pt}
	\caption{In the training process, the top figure illustrates the mAP evaluated on \texttt{minival} split at each epoch for different training scenarios. The bottom one shows the average IoUs between the clean (noise-free) box annotations and the ones available in these training scenarios of training samples.}
	\label{fig:process}
	\vspace{-10pt}
	\end{minipage}
\end{figure*}

\begin{equation}\label{equ:delta}
    \delta_i=f(IoU(\bm{b}_i,\bm{b}^*))\cdot \bm{p}_{i,c^*}.
\end{equation}
Here $\bm{p}_{i,c^*}$ represents the predicted classification score for category $c^*$, and $f(\cdot)$ should be an increasing function, and has several design choices. We can use step function ($f(t)=\mathbb{I}(t\geqslant \tau)$) or Gaussian function ($f(t)=e^{-\frac{(1-t)^2}{\alpha}}$) alternatively, where $\tau$ and $\alpha$ are the hyper-parameters as shown in Fig.~\ref{fig:f}. Both of them show comparably favorable performance in our experiments. The formulation in Equation~\eqref{equ:delta} makes sense for the reason that: 1) higher IoU between the manually annotated bounding box and predicted one indicates a higher probability that they refer to the same object; 2) higher classification score for the corresponding category roughly indicates higher confidence for the bounding box prediction. Moreover, higher IoU and score together contribute to higher confidence of $\bm{b}_i$ and hence higher $\delta_i$.

We will show the noise can be suppressed to a certain extent with the above prediction ensemble technique in Sec.~\ref{sec:PredictionensembleExperiment}.

\subsubsection{Teacher-Student Learning}\label{sec:TS}
Provided the above prediction ensemble method, a naive way to train a detector on the noisy dataset lies in applying such correcting method on the prediction obtained by the detector itself and utilizing the corrected annotations for supervision during training (denoted as \texttt{naive}). However, we experimentally find that it does not work to directly use this self-correction process in training. As shown in Fig.~\ref{fig:process}, although the corrected bounding boxes have higher IoUs than the clean ones than the original noisy annotations, the mAPs are not improved in the late of the process. We attribute it to the trivial solution obtained when the leading prediction dominates the ensemble. It is impracticable to count on the annotations corrected by the prediction itself to provide appropriate supervision. 

To this end, a Teacher-Student learning framework is adopted in our proposed training process. Specifically, a teacher detector is introduced to correct the annotations, which is pre-trained on the noisy dataset and kept frozen during the training process of the student detector (denoted as \texttt{TS}). The overall framework is shown in Fig.~\ref{fig:structure}. During training, the teacher and student detectors process the same images simultaneously. The predictions of the teacher detector are then utilized to correct the noisy bounding box annotations using our prediction ensemble method described in Sec.~\ref{sec:correct}. The corrected annotations are used as supervision for the student detector. In other words, the ability to alleviate noise of the teacher detector is transferred to the student detector, hence boosting the student detector's locating performance with no additional latency during the inference phase.

\section{Experiments}

\begin{table*}[h]
    \centering
    \caption{experimental results on COCO \texttt{test-dev} split.}
    \setlength{\tabcolsep}{10pt}
    \renewcommand{\arraystretch}{0.8} 
    \footnotesize
    \begin{tabular}{c|c|c|c|c|c|c c|c c c}
        \toprule
        Methods & type & scale & $\gamma$ & Methods & $AP$ & $AP_{50}$ & $AP_{75}$ & $AP_s$ & $AP_m$ & $AP_l$ \\
        \midrule
        \multirow{8}{*}{FCOS~\cite{FCOS}} & \multirow{5}{*}{synthesized} & \multirow{5}{*}{123k} & 0 & - & \textsl{38.9} & \textsl{57.9} & \textsl{42.0} & \textsl{22.4} & \textsl{41.4} & \textsl{48.6} \\
        \cline{4-11}
         & & & \multirow{2}{*}{0.05} & - & 37.1 & 57.4 & 40.2 & 21.4 & 39.3 & 46.0 \\
         & & & & Ours & \textbf{37.9 (+0.8)} & \textbf{58.1} & \textbf{41.3} & \textbf{22.1} & \textbf{40.6} & \textbf{47.0} \\
         \cline{4-11}
         & & & \multirow{2}{*}{0.1} & - & 33.6 & 55.4 & 36.0 & 19.6 & 35.8 & 41.4 \\
         & & & & Ours & \textbf{35.6 (+2.0)} & \textbf{56.9} & \textbf{38.8} & \textbf{20.8} & \textbf{38.0} & \textbf{44.2} \\
        \cline{2-11}
         & \multirow{3}{*}{real-world} & \multirow{3}{*}{12k} & 0 & - & \textsl{22.9} & \textsl{39.4} & \textsl{23.6} & \textsl{11.8} & \textsl{24.2} & \textsl{28.5} \\
        \cline{4-11}
         & & & \multirow{2}{*}{-} & - & 21.0 & 38.4 & 20.4 & 9.7 & 22.5 & 27.3\\
        & & & & Ours & \textbf{21.4 (+0.4)} & \textbf{38.8} & \textbf{21.0} & \textbf{10.2} & \textbf{23.0} & \textbf{27.4} \\
        \bottomrule
        \toprule
        \multirow{8}{*}{Faster R-CNN~\cite{FasterRCNN}} &  \multirow{5}{*}{synthesized} & \multirow{5}{*}{123k} & 0 & - & \textsl{37.8} & \textsl{58.9} & \textsl{41.0} & \textsl{22.0} & \textsl{40.7} & \textsl{46.8} \\
        \cline{4-11}
         & & & \multirow{2}{*}{0.05} & - & 36.4 & 58.5 & \textbf{39.7} & \textbf{21.5} & 39.4 & 44.5 \\
         & & & & Ours & \textbf{36.7 (+0.3)} & \textbf{59.0} & 39.4 & 21.2 & \textbf{39.7} & \textbf{45.2} \\
         \cline{4-11}
         & & & \multirow{2}{*}{0.1} & - & 33.7 & 57.1 & 35.9 & 20.1 & 36.8 & 41.2 \\
         & & & & Ours & \textbf{35.1 (+1.4)} & \textbf{58.1} & \textbf{37.6} & \textbf{20.4} & \textbf{38.2} & \textbf{43.2}\\
        \cline{2-11}
         & \multirow{3}{*}{real-world} & \multirow{3}{*}{12k} & 0 & - & \textsl{23.7} & \textsl{43.7} & \textsl{23.2} & \textsl{12.2} & \textsl{25.7} & \textsl{29.0} \\
        \cline{4-11}
         & & & \multirow{2}{*}{-} & - & 22.2 & 42.7 & 20.6 & 10.5 & 24.5 & 28.2 \\
        & & & & Ours & \textbf{22.5 (+0.3)} & \textbf{43.0} & \textbf{21.1} & \textbf{11.0} & \textbf{24.7} & \textbf{28.3} \\
        \bottomrule
    \end{tabular}
    \label{tab:results}
    \vspace{-10pt}
\end{table*}
We present experimental results on the COCO~\cite{coco} detection benchmark, which contains 118k images for training (\texttt{train} split), 5k images for validation (\texttt{minival} split) and 20k images for testing (\texttt{test-dev} split). We adopt the commonly used evaluation metric, AP, which is computed over ten different IoU thresholds, \ie, 0.5:0.05:0.95. Considering the rather high quality of the bounding box annotations in COCO, we regard the original annotations as the clean ones without man-made noise. Thus, we investigate our methods by training the models on the noisy dataset with either our newly-labeled or synthesized annotations and evaluating them on the original COCO \texttt{minival} split or \texttt{test-dev} split.

\subsection{Implementation Details}\label{sec:detail}

We benchmark on two representative detectors, FCOS~\cite{FCOS} and Faster R-CNN~\cite{FasterRCNN}, one of which is a single-stage method, and the other one follows the two-stage paradigm. All of our experiments are conducted based on mmdetection~\cite{mmdetection}, a popular open-source codebase, and adopt the default settings, except that we train FCOS\cite{FCOS} and Faster R-CNN~\cite{FasterRCNN} with a batch size of 8 and halve the learning rate according to linear scaling rule~\cite{linear-scaling-rule}, \ie, 0.005 for FCOS and 0.01 for Faster R-CNN. Unless specified, the models are trained for 12 epochs, with the learning rate decreased by 0.1 at 8 and 11 epochs. We use ResNet-50~\cite{resnet} as the default backbone and initialize it with the weights pre-trained on ImageNet~\cite{ImageNet}. FPN~\cite{fpn} is used by default. The input images are resized with the shorted side of 800 pixels and only horizontal image flipping is adopted as training data augmentations unless otherwise stated. 

\subsubsection{Training}

We adopt the teacher-student learning framework described in Sec.~\ref{sec:TS}. First, a teacher model is obtained by training on the dataset with noisy annotations. Then, the student model is trained. During training, the forward process of the teacher model is executed synchronously and the predictions along with the noisy annotations are fed into our proposed prediction ensemble module to produce the new bounding boxes for objects, which are then used as the ground truths to generate the supervision of the student detector. Following is a detailed description of how to get proper predictions from the teacher detector. For the FCOS, the predictions before NMS are utilized. Noting that the productions of classification scores and center-ness scores are considered as our category scores. As for the Faster R-CNN, the original one only produces sparse predictions after R-CNN, thus is not compatible with our methods. Nonetheless, a moderate modification is applied. Specifically, we collect the predictions of RPN and filter out the ones far away from any annotations (\eg, IoU $<$ 0.5) on the image plane or with rather low classification scores (\eg, less than 0.05). All the remaining predicted boxes are fed into the R-CNN as ROIs to perform the second-stage classification and regression. To avoid bringing a heavy computing burden in some extreme cases, only top-k scored predictions from RPN are collected and k is set to 1000 in our experiments. Then all these predictions obtained by R-CNN are utilized to correct the noisy annotations. Noting that the above strategies can be easily and smoothly extended to other one-/two-stage detectors, though we only instantiate one detector for each kind of them for the demonstration purpose.

\subsubsection{Inference}
Only the student detector is required during the inference phase, thus no additional computational cost is brought by our method. All the configurations in this phase can follow the default settings without additional modifications. Specifically, in post-processing, for FCOS, we adopt NMS with an IoU threshold set to 0.6 and a score threshold of 0.05. For Faster R-CNN, NMS with a 0.7 IoU threshold is applied after RPN and only the top 1000 boxes according to their classification scores are fed into R-CNN to get the final predictions. Finally, it is followed by another NMS with an IoU threshold set to 0.5 and a score threshold of 0.05.

\subsection{Overall results}

First, we evaluate our method on COCO \texttt{test-dev} split. All the models are trained on COCO \texttt{train} split with clean or noisy annotations. Different detectors (FCOS~\cite{FCOS} and Faster R-CNN~\cite{FasterRCNN}) are adopted as our baselines and the performance under the noise of various levels is presented in Table~\ref{tab:results}. It can be drawn that the performance is vulnerable to the noise in location annotations for both synthesized and real-world scenarios, while our method is capable to narrow the gap.
Furthermore, more improvement \wrt APs of higher thresholds is achieved by our methods, which indicates that our method tends to help the detectors to locate objects precisely. 

Besides, our method is only applied during the training phase, thus bringing no additional computational burdens for the inference phase. Considering only the forward pass is executed for the teacher detector and no intermediate results need to be cached, the runtime and memory occupation are also affected slightly in the training phase. Under our configuration, the training speed is about 80\% of the original one, and the memory occupation is about 1.02 times. The negligible burden makes our method easy for utilization and deployment. 

\subsection{Comparisons with Existing Methods}

In this section, we compare our method with some other existing methods. As stated in Sec.~\ref{sec:label-noise}, there are mainly two trends to deal with noisy labels. One focuses on reducing noisy samples' impact during training \cite{zhang2018generalized, arazo2019unsupervised, ghosh2017robust, li2019learning, Menon2020Can, jiang2018mentornet, ren2018learning, decoupling, coteaching}. However, these methods are specially designed for classification tasks and there are no publicly available works for adapting them to detection tasks. Thus, we make an effort to adapt one representative method among them, \ie, Co-teaching~\cite{coteaching}, to the detection task in our comparisons. Specifically, the two networks in the Co-teaching~\cite{coteaching} framework are replaced with two detectors (here, we adopt Faster R-CNN), and they are trained simultaneously with a shared backbone. Following the common practice, each network samples its anchors with small classification loss as the relatively clean instances. The regression loss in its peer network is only calculated on these sampled anchors while the others are ignored.

The other trend to deal with noisy labels focuses on correcting the noisy category labels \cite{reed2014training, xiao2015learning, tanaka2018joint, li2017learning} or noisy location annotations \cite{mao2020noisy}. However, the method in \cite{mao2020noisy} is only evaluated on the small Pascal VOC \cite{PascalVOC} dataset. Thus, we re-implement it under our setting as a strong counterpart.

The above two competitors are also trained on the COCO \texttt{train} split with our synthesized noisy annotations with $\gamma=0.1$. All other configurations follow those in Sec.~\ref{sec:detail}. The performance evaluated on the COCO \texttt{minival} split is presented in Table~\ref{tab:sota}. Our method achieves superior performance to the previous methods.

\begin{table}[t]
    \centering
    \caption{Comparisons with existing methods.}
    \setlength{\tabcolsep}{5.6pt}
    \renewcommand{\arraystretch}{1.0} 
    \footnotesize
    \begin{tabular}{c|c|cc|ccc}
        \hline
        Method & $AP$ & $AP_{50}$ & $AP_{75}$ & $AP_{s}$ & $AP_{m}$ & $AP_{l}$ \\
        \hline
        - & 33.3 & 56.3 & 35.6 & 19.3 & 37.5 & 42.1 \\
        Co-teaching \cite{coteaching} & 33.8 & 57.0 & 36.1 & 20.6 & 37.8 & 42.8 \\
        Mao \etal \cite{mao2020noisy} & 33.6 & 55.0 & 36.0 & 19.7 & 37.0 & 43.1 \\
        ours & \textbf{34.6} & \textbf{57.2} & \textbf{36.9} & \textbf{21.8} & \textbf{38.2} & \textbf{43.4} \\
        \hline
    \end{tabular}
    \label{tab:sota}
    \vspace{-10pt}
\end{table}

\subsection{Ablation Study}

In this section, we explore the effects of different components or design choices in our method. All the experiments are based on FCOS with synthesized Gaussian noise ($\gamma=0.1$).

\subsubsection{Prediction ensemble}\label{sec:PredictionensembleExperiment}

\begin{figure}[t]
	\centering  
	\includegraphics[width=0.85\linewidth]{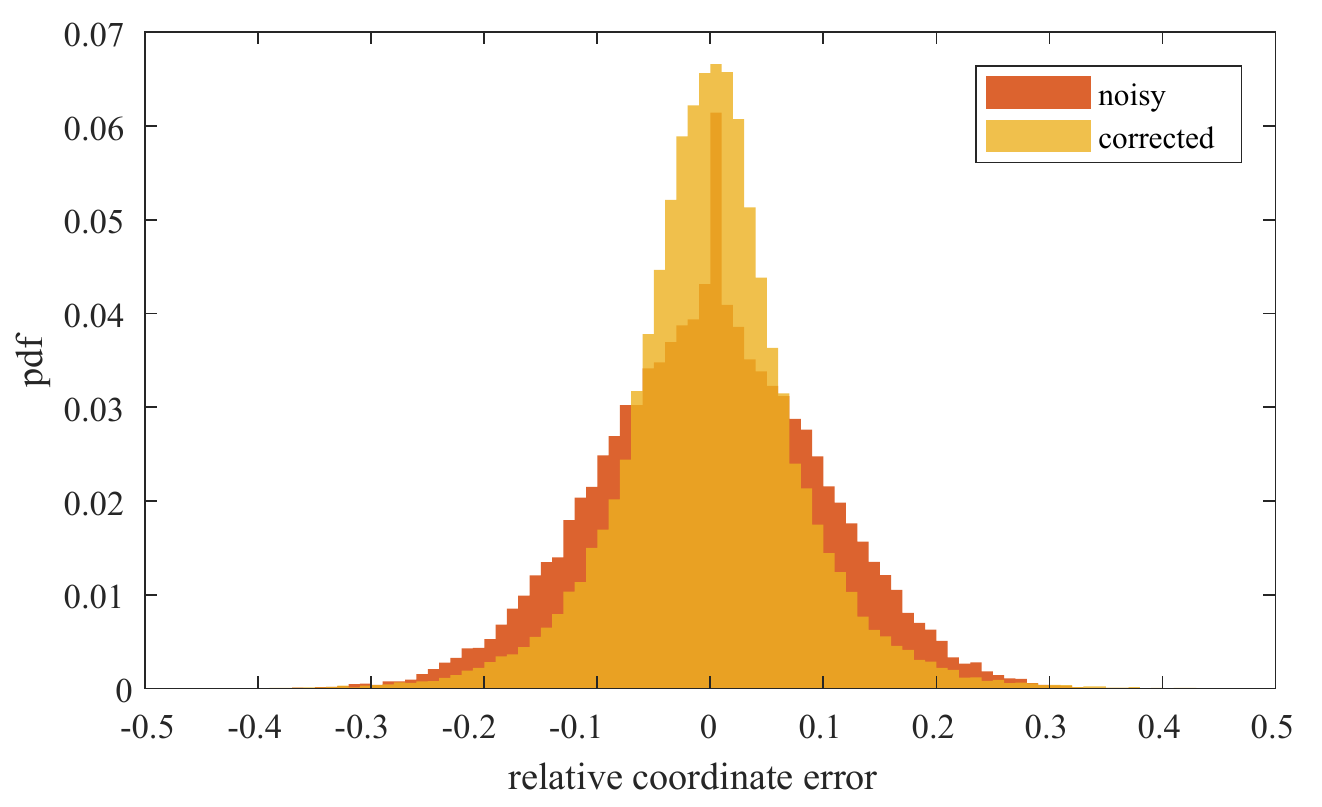}
	\vspace{-5pt}
	\caption{Distributions of relative boundary coordinate errors for noisy annotations and our corrected ones.}
	\label{fig:correct}
	\vspace{-5pt}
\end{figure}

As stated in Sec.~\ref{sec:correct}, our prediction ensemble component can correct the noise in location annotations effectively. We plot the distributions of relative boundary coordinate errors for noisy annotations and our corrected ones in Fig.~\ref{fig:correct}. The relative boundary coordinate errors are obtained by normalizing the absolute values of the difference between the noisy (or corrected) boundary coordinates and the accurate ones with respect to corresponding object widths or heights. Obviously, our corrected ones possess smaller variance, \ie, lower location errors, and thus contribute to better detection performance.

\begin{table}[t]
    \centering
    \caption{Different design choices of $\delta_i$ in Equation~\eqref{equ:delta}.}
    \setlength{\tabcolsep}{5.6pt}
    \renewcommand{\arraystretch}{0.9} 
    \footnotesize
    \begin{tabular}{c|c|c|c c}
        \hline
        $\delta_i$ & param. & $AP$ & $AP_{50}$ & $AP_{75}$  \\
        \hline
        $\bm{p}_{i,c^*}$ & - & 0.1 & 0.1 & 0.0  \\
        \hline
       $\mathbb{I}(IoU(\bm{b}_i,\bm{b}^*)\geqslant \tau)$ & $\tau=0.7$ & 34.9 & 56.3 & 37.7 \\
        \hline
       \multirow{3}{*}{$\mathbb{I}(IoU(\bm{b}_i,\bm{b}^*)\geqslant \tau)\cdot \bm{p}_{i,c^*}$} & $\tau=0.5$ & 34.6 & 57.2 & 36.9 \\
        & $\tau=0.6$ & 35.1 & \textbf{56.9} & \textbf{37.9} \\
        & $\tau=0.7$ & \textbf{35.2} & 56.6 & \textbf{37.9} \\
        \hline
       \multirow{3}{*}{$e^{-\frac{(1-IoU(\bm{b}_i,\bm{b}^*))^2}{\alpha}}\cdot \bm{p}_{i,c^*}$} & $\alpha=0.2$ & 35.0 & 56.3 & 37.7 \\
        & $\alpha=0.1$ & \textbf{35.2} & \textbf{57.0} & \textbf{38.1} \\
        & $\alpha=0.05$ & 33.3 & \textbf{57.0} & 34.7 \\
        \hline
    \end{tabular}
    \label{tab:delta}
    \vspace{-5pt}
\end{table}

\begin{table}[t]
    \centering
    \caption{Our method achieves consistent improvement under different training configurations. ``R50" and ``R101" are short for resnet50 and resnet101~\cite{resnet} respectively. ``MS" means training with multi-scale augmentation. ``2$\times$" is a training schedule that follows the setting explained in mmdetection~\cite{mmdetection}.}
    \setlength{\tabcolsep}{8pt}
    \renewcommand{\arraystretch}{0.9} 
    \footnotesize
    \begin{tabular}{c|c|c|c|c c}
        \hline
        Config. & $\gamma$ & Method & $AP$ & $AP_{50}$ & $AP_{75}$ \\
        \hline
        \multirow{5}{*}{R50+2$\times$+MS} & 0 & - & \textsl{40.9} & \textsl{59.8} & \textsl{44.1} \\
        \cline{2-6}
         & \multirow{2}{*}{0.05} & - & 38.8 & 58.6 & 42.2 \\
         & & ours & \textbf{39.2} & \textbf{59.8} & \textbf{43.2} \\
         \cline{2-6}
         & \multirow{2}{*}{0.1} & - & 35.1 & 57.3 & 38.0 \\
         & & ours & \textbf{36.2} & \textbf{57.9} & \textbf{39.3} \\
         \hline
         \hline
        \multirow{5}{*}{R101} & 0 & - & \textsl{40.7} & \textsl{59.6} & \textsl{43.9} \\
        \cline{2-6}
         & \multirow{2}{*}{0.05} & - & 39.0 & 59.3 & 42.8 \\
         & & ours & \textbf{39.8} & \textbf{59.9} & \textbf{43.5} \\
         \cline{2-6}
         & \multirow{2}{*}{0.1} & - & 35.7 & 58.1 & 38.3 \\
         & & ours & \textbf{37.3} & \textbf{59.0} & \textbf{40.5} \\
         \hline
    \end{tabular}
    \label{tab:config}
    \vspace{-5pt}
\end{table}

Then, we examine different design choices of $\delta_i$ in Equation~\eqref{equ:delta}, the default of which is $\delta_i=f(IoU(\bm{b}_i,\bm{b}^*))\cdot \bm{p}_{i,c^*}$. Without the former part related to IoU, we could not obtain meaningful results. It is anticipatable because numerous predictions not belonging to this object are involved in the ensemble and thus mislead the correcting procedure if the probability of the predictions corresponding to the object is excluded from consideration. In the case of removing the latter part ($\bm{p}_{i,c^*}$), slight performance degradation is observed, which confirms the effectiveness of using the scores. Moreover, we evaluate various functions for $f(\cdot)$, \eg, step function and Gaussian function. We observe that both functions could achieve comparable performance with appropriate hyper-parameters provided. We infer that functions with similar tendencies are workable with fine-tuned hyper-parameters. The overall experimental results are listed in Table~\ref{tab:delta}.

\subsubsection{Teacher-Student Learning}

In this section, we explore the essentials of our teacher-student learning framework. As discussed in Sec.~\ref{sec:TS}, the trivial solution is easy to obtain if we rely on the predictions produced by the detector itself to serve as the source of the ensemble for supervision. As shown in Fig.~\ref{fig:process}, for this \texttt{naive} approach, we can find that the quality of the annotations has consistent improvement over the original noisy annotations thanks to our self-correction method. Even at the very beginning of the training process, the quality is improved, which we attribute to the dynamic filtering ability of our designed $\delta_i$ (Equation~\eqref{equ:delta}). Great promotion on mAP is also witnessed for the first several epochs during training. However, the performance of the detector shows no further improvement over the one without correction for the last several epochs, though the IoU is still increasing. We may attribute it to that such an ensemble could not provide sufficient valuable correction for supervision in this stage. 

In our teacher-student learning framework (\texttt{TS}), the predictions used for correction and the ones under supervision are decoupled, and thus the trivial solution is avoided naturally. As shown in Fig.~\ref{fig:process}, the performance of the detectors is consistently improved during our whole training process. 

\subsection{Extensive Experiments}

\subsubsection{Other Training Configurations}

We examine the results of our method under other training configurations in this section as shown in Table~\ref{tab:config}. First, we test a longer training schedule with multi-scale training augmentation (R50+2$\times$+MS). Note that the configurations are adopted for both teacher and student training in our method. It can be found that even though better training configurations are adopted, the performance is still inferior when the location annotations are corrupted. But the gap can be narrowed by using our method as ever. A similar phenomenon is observed in the case of a larger backbone (R101). 

Then we test the case with the training configurations of the teacher and student detectors decoupled, \eg, a stronger teacher is obtained first and then utilized to correct the supervision of the student detector. The experimental results are shown in Table~\ref{tab:tc-sc}. First, we compare with the iterative training strategy, \ie, the student detector obtained from our method is then used as a teacher for the second round of training in our framework. However, we observe no continuous improvement, which we attribute to too much inductive bias involved in iterative training, leading to accumulated errors. Then we test the stronger teachers obtained by the longer training schedule with multi-scale training augmentation (R50+2$\times$+MS) or a larger backbone (R101). We observe that stronger teachers lead to better student performance, but the additional improvement is limited. An upper bound is likely to exist, which prevents the performance from approximating to the optimal result infinitely. 

\begin{table}[t]
    \centering
    \setlength{\tabcolsep}{11pt}
    \caption{Training with the decoupled teacher and student configurations. The rows with ``S Config." marked as ``-" show the results of isolated teacher detectors. The rows with ``T ~Config." marked as ``-" present the results obtained without our correction method and teacher-student learning, which are also our baselines for each group. The rows with ``T ~Config." marked as ``R50*" are for the comparison with the iterative training strategy. Better performance is achieved when stronger teachers are adopted.}
    \vspace{-5pt}
    \setlength{\tabcolsep}{15pt}
    \renewcommand{\arraystretch}{0.9} 
    \footnotesize
    \begin{tabular}{c|c|c|c}
    \hline
        S Config. & T Config. & $\gamma$ & $AP$ \\
        \hline
        - & R50+2$\times$+MS & 0 & \textsl{40.9} \\
        \hline
        - & R101 & 0 & \textsl{40.7} \\
        \hline
        \hline
        \multirow{11}{*}{R50} & - & 0 & \textsl{38.5} \\
        \cline{2-4}
         & - & \multirow{5}{*}{0.05} & 37.1 \\
         & R50 & & 37.7 (+0.6)\\
         & R50* & & 37.5 (+0.4)\\
         & R50+2$\times$+MS & & 37.8 (+0.7)\\
         & R101 & & \textbf{37.9 (+0.8)} \\
         \cline{2-4}
         & - & \multirow{5}{*}{0.1} & 33.5 \\
         & R50 & & 35.2 (+1.7)\\
         & R50* & & 35.1 (+1.6)\\
         & R50+2$\times$+MS & & 35.4 (+1.9)\\
         & R101 & & \textbf{35.5 (+2.0)} \\
        \hline
    \end{tabular}
    \label{tab:tc-sc}
    \vspace{-5pt}
\end{table}

\subsubsection{Robustness}

\begin{table}[t]
    \centering
    \caption{Robustness test with 3 runs for each configuration.}
    \resizebox{\linewidth}{!}{
    \begin{tabular}{l c|c|c|c}
        \hline
         \multicolumn{2}{c|}{\multirow{2}{*}{Config}} & \multicolumn{3}{c}{mAP} \\
        \cline{3-5}
         & & $\gamma$=0 & $\gamma$=0.05 & $\gamma$=0.1 \\
        \hline
        a. & Same annotations without noise & 38.5($\pm$0.0) & - & - \\
        \hline
        b. & annotations with the same noise & - & 37.1($\pm$0.1) & 33.7($\pm$0.2) \\
        c. & + our methods & - & 37.7($\pm$0.1) & 35.3($\pm$0.1)\\
        \hline
        d. & annotations with different noise & - & 37.1($\pm$0.2) & 33.4($\pm$0.2) \\
        e. & + our methods & - & 37.7($\pm$0.1) & 35.1($\pm$0.1)\\
        \hline
    \end{tabular}
    }
    \label{tab:robustness}
    \vspace{-5pt}
\end{table}

As randomness is involved in our noise generation, the robustness of our method should be taken into account. So we run 3 times for each configuration to test it, as shown in Table~\ref{tab:robustness}. First, we conduct experiments on datasets with the same clean (a) or noisy (b and c) annotations for each run. In this case, the randomness comes from the random initialization of the model weights and the random sampling order of the data loader. We find that the performance varies much more if the annotations are noisy (compare b to a). Then, we execute noise generation 3 times resulting in 3 different sets of noisy annotations for the dataset (d and e). This time, the performance possesses a larger variance due to the additional source of randomness, \ie, noise in the annotations (compare d to b). Furthermore, our method achieves consistent gains across different settings (compare c to b and e to d), which demonstrates the robustness of our method. 

\subsection{Label Time Analysis}\label{sec:real}
In this section, we analyze the time consumption of labeling the images. It costs us about 364 human hours to label 12k images with rough bounding box annotations. Nevertheless, we can only label about 8k images with fine-grained bounding box annotations by taking up nearly the same time. That is because much more time is consumed to finely adjust the boundary annotations. Thus, about 1.5 times faster labeling speed can be achieved by only requiring rough bounding box annotation. We believe that much labeling time can be saved if we further loosen the label restrictions. With the 8k images with elaborate labels, we can train a detector as shown in Table~\ref{tab:real}. Its performance is far behind those trained with 12k images, especially for $AP_{50}$ (up to $-4.2$ gap), though they are labeled with almost equal time consumed. While by leveraging our method, we can narrow the gap for $AP_{50}$ to only $-0.3$. We may conclude that, with a limited or restricted label budget, it is preferable to label with rough box annotations rather than elaborate but fewer ones. 
\begin{table}[t]
    \caption{Experimental results for detectors trained with our newly labeled subset of COCO. }
    \vspace{-5pt}
    \setlength{\tabcolsep}{8pt}
    \renewcommand{\arraystretch}{0.9} 
    \footnotesize
    \centering
    \begin{tabular}{l|c|c|c|c c}
        \hline
        dataset \& methods & scale & time & $AP$ & $AP_{50}$ & $AP_{75}$ \\
        \hline
        COCO labels & 12k & - & \textsl{22.8} & \textsl{39.2} & \textsl{23.3} \\
        \hline
        our noisy labels & \multirow{2}{*}{12k} & \multirow{2}{*}{364h} & 21.2 & 38.4 & 20.7 \\
        \quad + our method &  &  & \textbf{21.5} & \textbf{38.9} & \textbf{21.2} \\
        \hline
        elaborate labels & 8k & 372h & 19.8 & 35.0 & 19.8 \\
        \hline
    \end{tabular}
    \label{tab:real}
    \vspace{-5pt}
\end{table}

\subsection{Evaluation on Other Forms of Noise}
Finally, we introduce two additional forms of noisy location annotations for analyzing their impact on the performance degradation. We believe this can further provide positive guidance for data annotation. Specifically, we adopt asymmetrical noise, one of which restricts the noisy annotation box to completely enclose the ideal one and the other completely enclosed by the ideal one. Exponential distributions with hyper-parameter $\lambda$ are adopted. We set $\lambda=20\sqrt{2}$, and thus the noise level is $\gamma=0.05$. We find that the asymmetrical noise has a rather more severe impact on the detectors' performance degradation than the symmetrical one (Gaussian distribution) as shown in Table~\ref{tab:asym}, with APs of higher IoU thresholds dropping devastatingly. While our method can still reduce the negative impact to a certain extent. Based on the above observations, we conclude that it is better to control the distribution of the annotated coordinates centered as near as possible to the ideal one, and neither consistently larger nor smaller noisy annotations than the ideal one are better choices for box labeling.

\begin{table}[t]
    \centering
    \caption{Asymmetrical noise has rather more terrible impact on the detectors' performance than the symmetrical ones.}
    \vspace{-5pt}
    \setlength{\tabcolsep}{9.5pt}
    \renewcommand{\arraystretch}{0.9} 
    \footnotesize
    \begin{tabular}{c|c|c|cc}
        \hline
        Noise & Method & $AP$ & $AP_{50}$ & $AP_{75}$\\
        \hline
        - & - & 38.5 & 57.3 & 41.6\\
        \hline
        Sym. ($\gamma=0.05$) & - & 37.1 & 56.9 & 40.6\\
        \hline
        Asym. & - & 34.6 & 56.6 & 37.9\\
        ($\gamma=0.05$, enclosing) & ours & 35.1 & 56.6 & 38.8\\
         \hline
        Asym. & - & 34.1 & 56.4 & 37.4\\
        ($\gamma=0.05$, enclosed) & ours & 34.5 & 56.4 & 38.0\\
        \hline
    \end{tabular}
    \label{tab:asym}
    \vspace{-5pt}
\end{table}
\section{Conclusions}
The impact of noisy location annotations for detection tasks is investigated in the paper. We demonstrate that fine-grained location annotations are crucial for the plausible performance of modern detectors, especially for the AP performance with high IoU thresholds. Then a simple yet effective method is proposed to correct the noise in the location annotations, which can remarkably improve the performance of detectors under both synthesized noisy annotations and real-world man-made noisy annotations. Besides, we find that if there is a limited or restricted label budget, the rough location annotations with some weak noise can significantly reduce label cost, and utilizing our method is essential for performance improvement. Furthermore, in our view, more studies should be focused on the pattern of man-made noise, based on which much more effective methods could be proposed and further improve the detector performance. Finally, the combination of location noise and category label noise is also a challenging topic, which needs future exploration.

\section*{Acknowledgment}
The authors would like to acknowledge the data label team members, for their effort to construct the new real-world noisy detection dataset.

\ifCLASSOPTIONcaptionsoff
  \newpage
\fi

\clearpage



\bibliographystyle{IEEEtran}
\bibliography{bib}
\end{document}